\documentclass{article} 
\usepackage{nips12submit_e,times}
\usepackage{graphicx}
\usepackage{natbib}

\nipsfinalcopy

\title{Big Neural Networks Waste Capacity}

\author{
Yann N. Dauphin \& Yoshua Bengio\\
D\'epartement d'informatique et de recherche op\'erationnelle\\
    Universit\'e de Montr\'eal,  Montr\'eal, QC, Canada
}

\begin{document}

\maketitle

\begin{abstract}
This article exposes the failure of some big neural networks to leverage added capacity
to reduce underfitting.
Past research suggest diminishing returns when increasing the size
of neural networks. Our experiments on ImageNet LSVRC-2010 show
that this may be due to the fact there are highly diminishing returns for
capacity in terms of training error, leading to underfitting. This suggests that the
optimization method - first order gradient descent - fails at this regime.
Directly attacking this problem, either through the optimization method or the
choices of parametrization, may allow to improve the generalization error on 
large datasets, for which a large capacity is required.
\end{abstract}

\section{Introduction}

Deep learning and neural networks have achieved state-of-the-art results on
vision~\footnote{\citet{Krizhevsky-2012} reduced by almost one half the
  error rate on the 1000-class ImageNet object recognition benchmark},
language~\footnote{\citet{Mikolov-Interspeech-2011} reduced perplexity
on WSJ by 40\% and speech recognition absolute word error rate by $>1$\%.}
, and audio-processing tasks~\footnote{For speech
  recognition,~\citet{Seide2011} report relative word error rates
  decreasing by about 30\% on datasets of 309 hours}. All these cases
involved fairly large datasets, but in all these cases, even larger ones
could be used. One of the major
challenges remains to extend neural networks on a much larger scale, and with
this objective in mind, this paper asks a simple question: is there an
optimization issue that prevents efficiently training larger networks?

Prior evidence of the failure of big networks in the litterature can be found 
for example in \citet{Coates2011}, which shows that increasing the capacity of certain neural net
methods quickly reaches a point of diminishing returns on the test error. These
results have since been extended to other types of auto-encoders and RBMs
\citep{Salah+al-2011,Courville+al-2011}. Furthermore, \cite{Coates2011} shows
that while neural net methods fail to leverage added capacity K-Means can.
This has allowed K-Means to reach state-of-the-art performance on CIFAR-10
for methods that do not use artificial transformations. This is an unexpected
result because K-Means is a much \emph{dumber} unsupervised learning algorithm
when compared with RBMs and regularized auto-encoders. \cite{Coates2011} argues that this
is mainly due to K-Means making better use of added capacity, but it does
not explain why the neural net methods failed to do this.

\section{Experimental Setup}

We will perform experiments with the well known ImageNet LSVRC-2010
object detection dataset\footnote{http://www.image-net.org/challenges/LSVRC/2010/}. The subset used in the \emph{Large
Scale Visual Recognition Challenge 2010} contains 1000 object categories
and 1.2 million training images.

This dataset has many attractive features:
\begin{enumerate}
    \item The task is difficult enough for current algorithms that there is
    still room for much improvement. For instance, \cite{Krizhevsky-2012} was
    able to reduce the error by half recently. What's more the state-of-the-art
    is at 15.3\% error. Assuming minimal error in the human labelling of the dataset,
    it should be possible to reach errors close to 0\%.
    \item Improvements on ImageNet are thought to be a good proxy for progress
    in object recognition \citep{imagenet_cvpr09}.
    \item It has a large number of examples. This is the setting that is
    commonly found in industry where datasets reach billions of examples.
    Interestingly, as you increase the amount of data, the
    training error converges to the generalization error. In other
    words, reducing
    training error is well correlated with reducing generalization error, 
    when large datasets are available.
    Therefore, it stands to reason that resolving underfitting problems may
    yield significant improvements.
\end{enumerate}

We use the features provided by the \emph{Large
Scale Visual Recognition Challenge 2010}\footnote{http://www.image-net.org/challenges/LSVRC/2010/download-public}. The images are convolved with SIFT features, then K-Means is used to
form a visual vocabulary of 1000 visual words. Following the litterature, we
report the Top-5 error rate only.

The experiments focus on the behavior of Multi-Layer Perceptrons (MLP) as
capacity is increased. This is done by increasing the number of hidden units
in the network. The final classification layer of the network is a softmax
over possible classes ($softmax({\bf x}) = {e^{-{\bf x}}}/{\sum_i e^{-{\bf x}_i}}$).
The hidden layers use
the logistic sigmoid activation function ($\sigma({\bf x}) = \;^{1}/_{(1 + e^{-{\bf x}})}$).
We initialize the weights of the hidden layer according to the formula
proposed by \cite{GlorotAISTATS2010}. The parameters of the classification
layer are  initialized to 0, along with all the bias (offset) parameters of the MLP.

The hyper-parameters to tune are the learning rate and the number of hidden
units. We are interested in optimization performance so we cross-validate
them based on the training error. We use a grid search with the learning
rates taken from $\{ 0.1, 0.01 \}$ and the number of hiddens from
$\{ 1000, 2000, 5000, 7000, 10000, 15000 \}$. When we report the
performance of a network with a given number of units we choose the best learning
rate. The learning rate is decreased by 5\% everytime the training error goes up
aftern an epoch. We do not use any regularization because it would typically not
help to decrease the training set error. The number of epochs is set to 300 s
that it is large enough for the networks to converge.

The experiments are run on a cluster of Nvidia Geforce GTX 580 GPUs with the
help of the Theano library \citep{bergstra+al:2010-scipy}. We make use of HDF5
\citep{folk2011overview} to load
the dataset in a lazy fashion because of its large size.
The shortest training experiment took 10 hours to run and
the longest took 28 hours.

\section{Experimental Results}

\begin{figure*}[h]
\centering
\includegraphics[width=.8\textwidth]{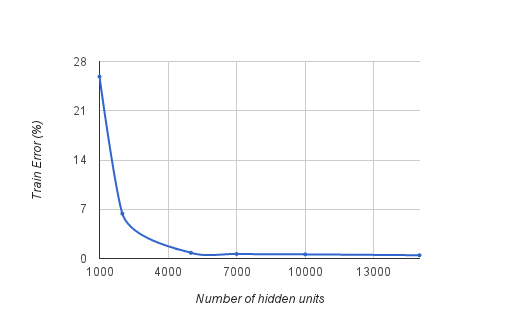}
\caption{Training error with respect to the capacity of a 1-layer sigmoidal
neural network. This curve seems to suggest we are correctly leveraging
added capacity.}
\label{fig:capacity}
\end{figure*}

Figure \ref{fig:capacity} shows the evolution of the training error as the
capacity is increased. The common intuition is that this increased capacity will
help fit the training set - possibly to the detriment of generalization error.
For this reason practitioners have focused mainly on the problem of overfitting
the dataset when dealing with large networks - not underfitting. In fact, much
research is concerned with proper regularization of such large networks
\citep{Hinton-et-al-arxiv2012,Hinton06}.

\begin{figure*}[h]
\centering
\includegraphics[width=.8\textwidth]{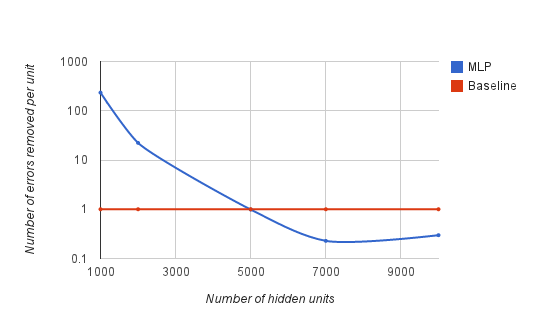}
\caption{Return on investment on the addition of hidden units for a 1-hidden layer sigmoidal neural network.
The vertical axis is the number of training errors removed per additional hidden unit, after 300 epochs. We see
here that it is harder and harder to use added capacity.}
\label{fig:roi}
\end{figure*}

However, Figure \ref{fig:roi} reveals a problem in the training of big networks.
This figure is the derivative of the curve in Figure \ref{fig:capacity} (using
the number of errors instead of the percentage). It may
be interpreted as the return on investment (ROI) for the addition of capacity. The
Figure shows that the return on investment of additional hidden units decreases fast,
where in fact we would like it to be close to constant. Increasing the capacity
from 1000 to 2000 units, the ROI decreases by an order of magnitude.
It is harder and harder for the model to make use of additional capacity. 
The red line is a baseline where the additional unit is used as a template
matcher for one of the training errors. In this case, the number of errors
reduced per unit is at least 1. We see that the MLP does not manage to beat this
baseline after 5000 units, {\em for the given number of training iterations}.

\begin{figure*}[h]
\centering
\includegraphics[width=.9\textwidth]{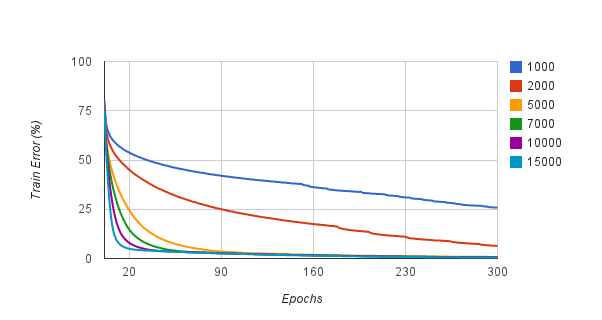}
\caption{Training error with respect to the number of epochs of gradient descent. Each line is a 1-hidden layer sigmoidal neural network with a different number of hidden units.}
\label{fig:epochs}
\end{figure*}

For reference, we also include the learning curves of the networks used for Figure
\ref{fig:capacity} and \ref{fig:roi} in Figure \ref{fig:epochs}. We see that the curves for
capacities above 5000 all converge towards the same point.

\section{Future directions}

This rapidly decreasing return on investment for capacity in big networks
seems to be a failure of first order gradient descent.

In fact, we know that the first order
approximation fails when there are a lot of interactions between hidden units.
It may be that adding units increases the interactions between units and causes
the Hessian to be ill-conditioned. This reasoning suggests two research directions:
\begin{itemize}
    \item methods that break interactions between large numbers of units. This
    helps the Hessian to be better conditioned and will lead to better
    effectiveness for first-order descent.
    This type of method can be implemented efficiently. Examples of this approach
    are sparsity and orthognality penalties.
    \item methods that model interactions between hidden units. For example,
    second order methods \citep{martens2010hessian} and natural gradient
    methods \citep{NIPS2007-56}. Typically, these are expensive approaches
    and the challenge is in scaling them to large datasets, where stochastic
    gradient approaches may dominate. The ideal target is a stochastic natural
    gradient or stochastic second-order method.
\end{itemize}
The optimization failure may also be due to other reasons. For example, networks
with more capacity have more local minima. Future work should investigate tests
that help discriminate between ill-conditioning issues and local minima issues.

Fixing this optimization problem may be the key to unlocking better performance
of deep networks. Based on past observations~\citep{Bengio-2009-book,Erhan+al-2010-small}, 
we expect this optimization
problem to worsen for deeper networks, and our experimental setup should be extended
to measure the effect of depth.
As we have noted earlier, improvements on the training set error should be well
correlated with generalization for large datasets.

\bibliography{strings,ml,aigaion,paper}
\bibliographystyle{natbib}

\end{document}